\documentclass[runningheads]{llncs}
\usepackage[utf8]{inputenc}
\usepackage{cite}
\usepackage{graphicx}
\usepackage{comment}
\usepackage{amsmath,amssymb} 
\usepackage{color}
\usepackage{xcolor}
\usepackage{microtype}
\usepackage[position=bottom]{subfig}

\usepackage{gensymb}
\usepackage{array}
\usepackage{booktabs}
\usepackage{makecell}
\usepackage{multirow}
\usepackage[font=small,labelfont=bf,tableposition=top]{caption}
\usepackage{arydshln}
\newcolumntype{?}{!{\vrule width 1pt}}
\newcommand{\ra}[1]{\renewcommand{\arraystretch}{#1}}

\newcommand\gray[1]{\textcolor{gray}{#1}}
\newcommand{\code}{\texttt}
\usepackage{soul}

\definecolor{orange}{rgb}{1.0,0.5,0}
\definecolor{DarkGreen}{rgb}{0,0.5,0}

\definecolor{pink}{RGB}{255,192,203}
\definecolor{paleturquoise}{RGB}{175,238,238}
\definecolor{lightsteelblue}{RGB}{176,196,222}
\definecolor{violet}{RGB}{238,130,238}
\definecolor{orange}{RGB}{255,120,0}
\definecolor{mediumpurple}{RGB}{147,112,219}
\definecolor{lightgreen}{RGB}{0,128,0}

\usepackage[pagebackref=true,breaklinks=true,urlcolor=blue,letterpaper=true,colorlinks,bookmarks=false]{hyperref}

\def\etal{\emph{et al.}}

\def\m#1{\ensuremath{\mathtt{#1}}}

\def\v#1{\ensuremath{\mathbf{#1}}}

\def\I{\m{I}}
\def\H{\m{H}}

\def\A{\text {A}}
\def\B{\text {B}}
\def\p{\v{p}}
\def\q{\v{q}}
\def\C{\m{C}}

\begin{document}
\pagestyle{headings}
\mainmatter
\def\ECCVSubNumber{997}  

\title{S2DNet: Learning Accurate Correspondences for Sparse-to-Dense Feature Matching}

\titlerunning{S2DNet: Learning Accurate Correspondences for S2D Feature Matching}
\author{
  Hugo Germain\inst{1} \and Guillaume Bourmaud\inst{2} \and Vincent Lepetit\inst{3}}
\institute{
  LIGM, 
  École des Ponts, Univ Gustave Eiffel, 
  CNRS, Marne-la-vallée, France \and
  Laboratoire IMS, Université de Bordeaux, France}

\maketitle


\begin{abstract}
  Establishing robust and accurate correspondences  is a fundamental backbone to
  many computer vision algorithms.  While recent learning-based feature matching
  methods have shown promising results in providing robust correspondences under
  challenging conditions, they are often limited in terms of precision.  In this
  paper, we  introduce S2DNet, a  novel feature matching pipeline,  designed and
  trained to efficiently establish both  robust and accurate correspondences. By
  leveraging  a sparse-to-dense  matching paradigm,  we cast  the correspondence
  learning problem as a supervised classification task to learn to output highly
  peaked  correspondence maps.   We show  that S2DNet  achieves state-of-the-art
  results  on the  HPatches  benchmark, as well  as  on several
  long-term visual localization datasets. 

  
  
\end{abstract}



\section{Introduction}



Establishing both 
accurate and robust correspondences across images is an
underpinning step to many computer vision algorithms, such as
Structure-from-Motion (SfM)~\cite{heinly2015_reconstructing_the_world,
schoenberger2016sfm, schoenberger2016mvs, sweeney2016}, visual tracking~\cite{
visualtracking, visualtracking2} and visual localization~\cite{CSL,
Svrm2014AccurateLA, Sattler:hal-01513083}. Yet, obtaining such correspondences
in long-term scenarios where extreme visual changes can appear remains an
unsolved problem, as shown by recent benchmarks~\cite{6DOFBenchmark,
Taira2018InLocIV}. In particular, illumination (e.g. daytime to nighttime),
cross-seasonal and structural changes are very challenging factors for feature matching.

\begin{figure}[hbtp] \centering
  \raisebox{-0.35\height}{\includegraphics[width=0.2\textwidth]{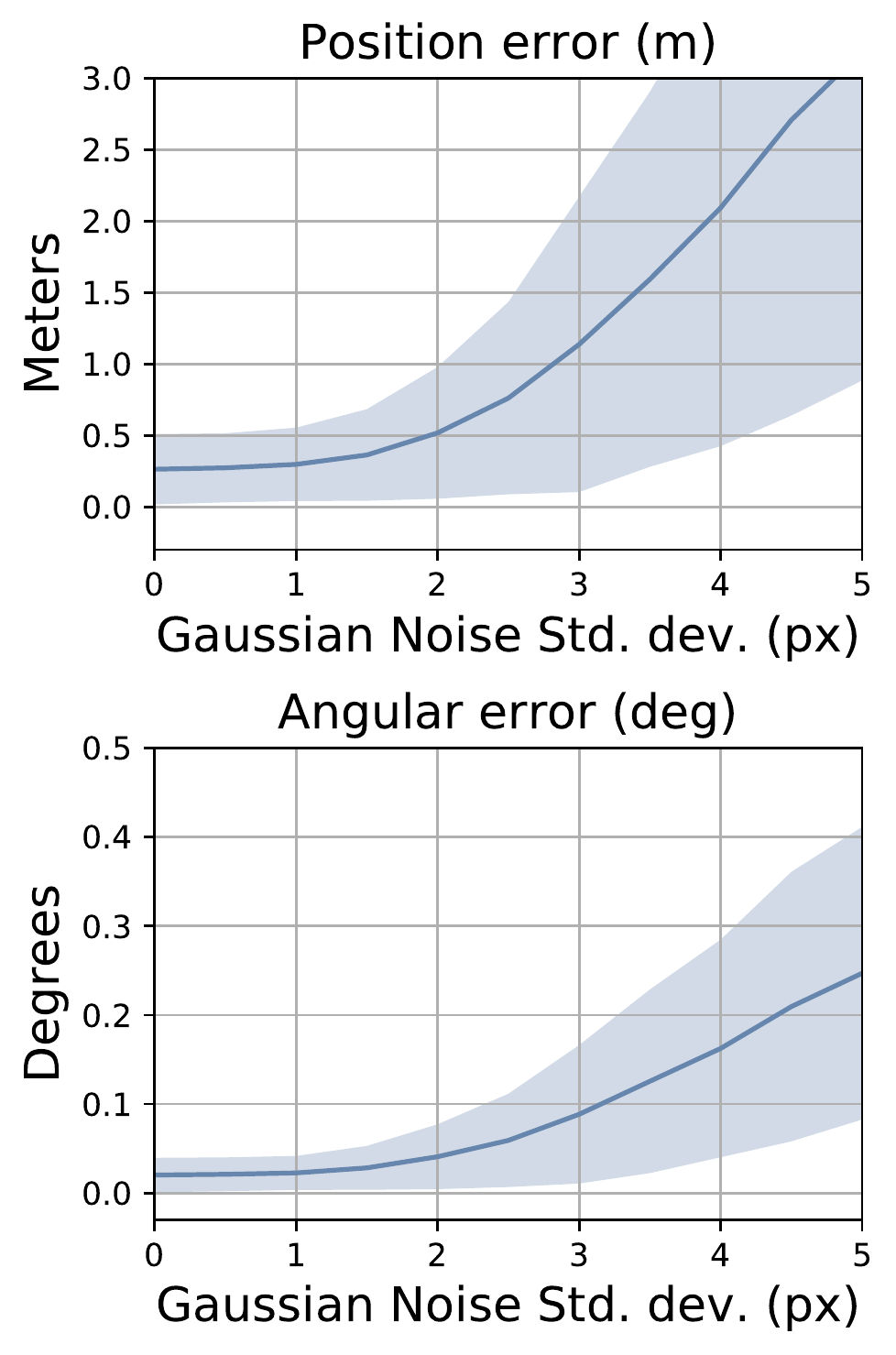}}
  \raisebox{-0.4\height}{\includegraphics[width=0.7\textwidth]{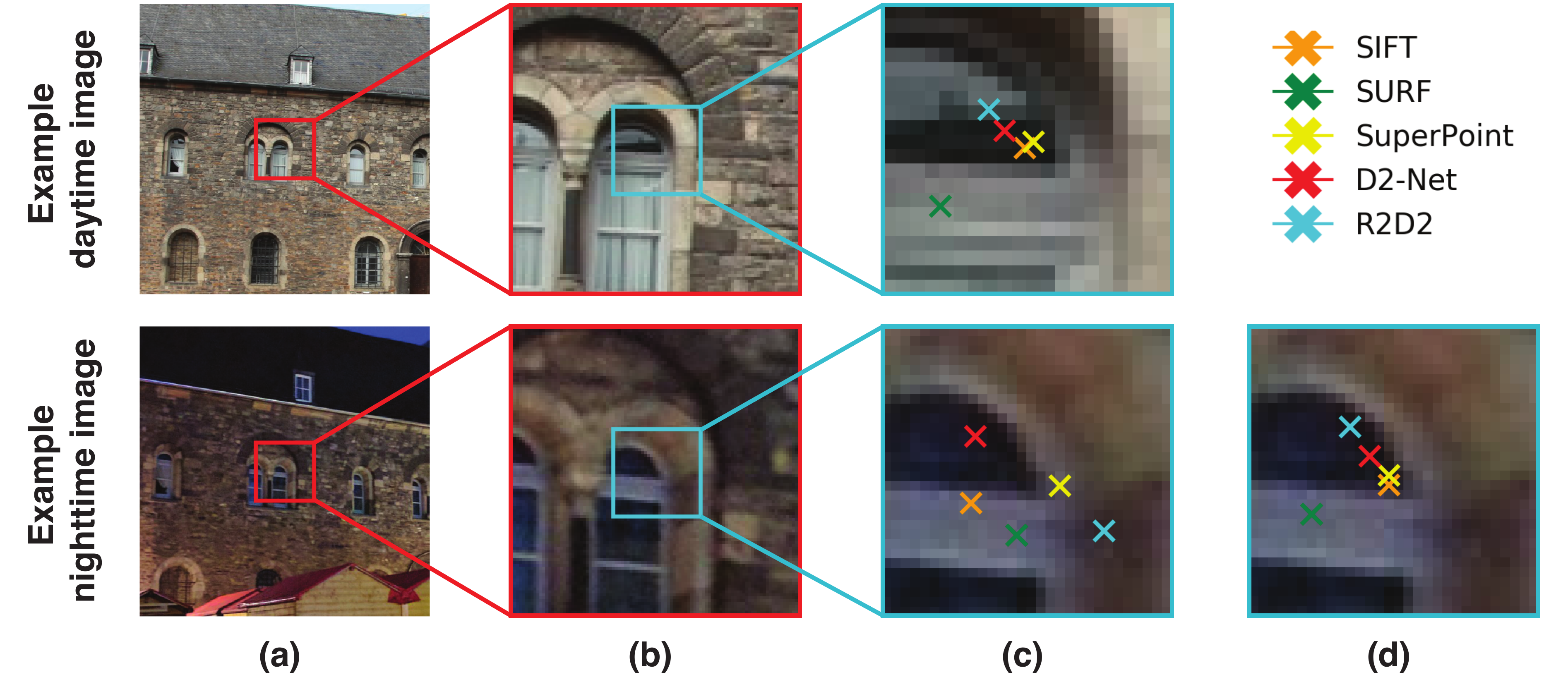}}
  \caption{\textbf{Learning accurate correspondences.}  On the left, we report
    the impact of adding a gaussian noise of increasing variance on ground-truth
    2D-3D correspondences   for   the   task   of   visual   localization,   on
    Aachen Day-Night~\cite{Sattler2012ImageRF,6DOFBenchmark}  images.    This
    experiment highlights  the importance  of having very  accurate
    correspondences,  as offsets of a few  pixels can  lead to
    localization errors  of several meters.  Yet as shown on the right,
    sparse-to-sparse methods  fail to make such  accurate predictions. We show
    in  $(a)$ and $(b)$ local regions of interest for a day-night image pair. In
    $(c)$ [top], we display the keypoint detections being the nearest to the
    center of the  patch in  the daytime  image for each detector;  [bottom] we
    show the closest correspondent detected keypoints for  each detector in the
    nighttime image.
    In  $(d)$, we  show  the correspondent  image locations found by S2DNet in
  the nighttime image for daytime keypoint detections.  S2DNet  manages  to
find   much  more  accurate correspondences than sparse-to-sparse methods.}
\label{fig:intro_figure} \end{figure}

The correspondences accuracy plays a major role in the performance of the
aforementioned algorithms. Indeed, the noise perturbation experiment
of Figure~\ref{fig:intro_figure} (left) shows the highly
damaging impacts of errors of a few pixels on visual localization.
A traditional and very commonly used paradigm for feature matching between two images
consists in detecting
a set of keypoints~\cite{affinedetectors, affinedetectors2,
Harris88acombined, SURF, superpoint, SIFT,
Savinov2016QuadNetworksUL, LIFT,D2Net,R2D2}, followed by a description
stage~\cite{Balntas2016LearningLF, Brown2011DiscriminativeLO,
Dong2014DomainsizePI, SimoSerra2015DiscriminativeLO, Simonyan2014LearningLF,
LFNet, superpoint, D2Net, R2D2, LIFT, SIFT, DELF} in each image. Sparse
sets of keypoints and their descriptors are then matched using for
instance approximate nearest neighbours. This \textit{sparse-to-sparse} matching
approach has the main advantage of being both computationally and memory efficient.
For long-term scenarios, this requires detecting repeatable keypoints and computing
robust descriptors, which is very challenging.

With the advent of convolutional neural networks~(CNNs), learning-based
sparse-to-sparse matching methods have emerged, attempting to
improve robustness of both detection and description stages in an
end-to-end fashion. Several single-CNN pipelines~\cite{superpoint, D2Net, R2D2}
were trained with pixel-level supervision to jointly detect \textit{and}
describe interest points. These methods have yielded very competitive
results especially in terms of number of correct matches, but fail to deliver
highly accurate correspondences~\cite{D2Net, R2D2}.
Indeed detecting the same keypoints repeatably across images is very challenging
under strong visual changes, as illustrated in Figure~\ref{fig:intro_figure}
(right). Thus, the accuracy of such methods becomes highly reliant on the feature detector's
precision and repeatability.


A recently proposed alternative~\cite{3DV2019} to solve the repeatable keypoint
detection issue is to shift the sparse-to-sparse paradigm into a 
\textit{sparse-to-dense} approach: Instead of trying to
detect consistent interest points across images, feature detection is 
performed asymmetrically and correspondences are searched exhaustively in the other image.
That way, all the information in the challenging image is preserved, allowing
each pixel to be a potential correspondent candidate. Thanks to the
popularization and development of GPUs, this exhaustive search can be done
with a small computational overhead compared to other learning-based methods.
%
This approach has showed to give competitive results~\cite{3DV2019} 
when trained in a weakly supervised fashion for the task of image retrieval.

In this paper, we reuse the sparse-to-dense idea while also addressing the
accuracy issue. We introduce a novel feature matching pipeline which we
name \textit{S2DNet}. We explicitly designed it to learn both robust and
accurate correspondence maps. Instead of losses operating at a sparse descriptor level,
we train S2DNet to learn both accurate and discriminative correspondence
maps. By casting the feature matching problem as a classification problem,
we give rich feedback on correspondences errors at training time. Compared to other
approaches, we show that our pipeline generates much more accurate
correspondences with most off-the-shelf feature detectors.
Our evaluations show that we achieve state-of-the-art results on
image matching and long-term visual localization benchmarks.


\begin{figure}[t]
\begin{center}
  \includegraphics[width=1.0\columnwidth]{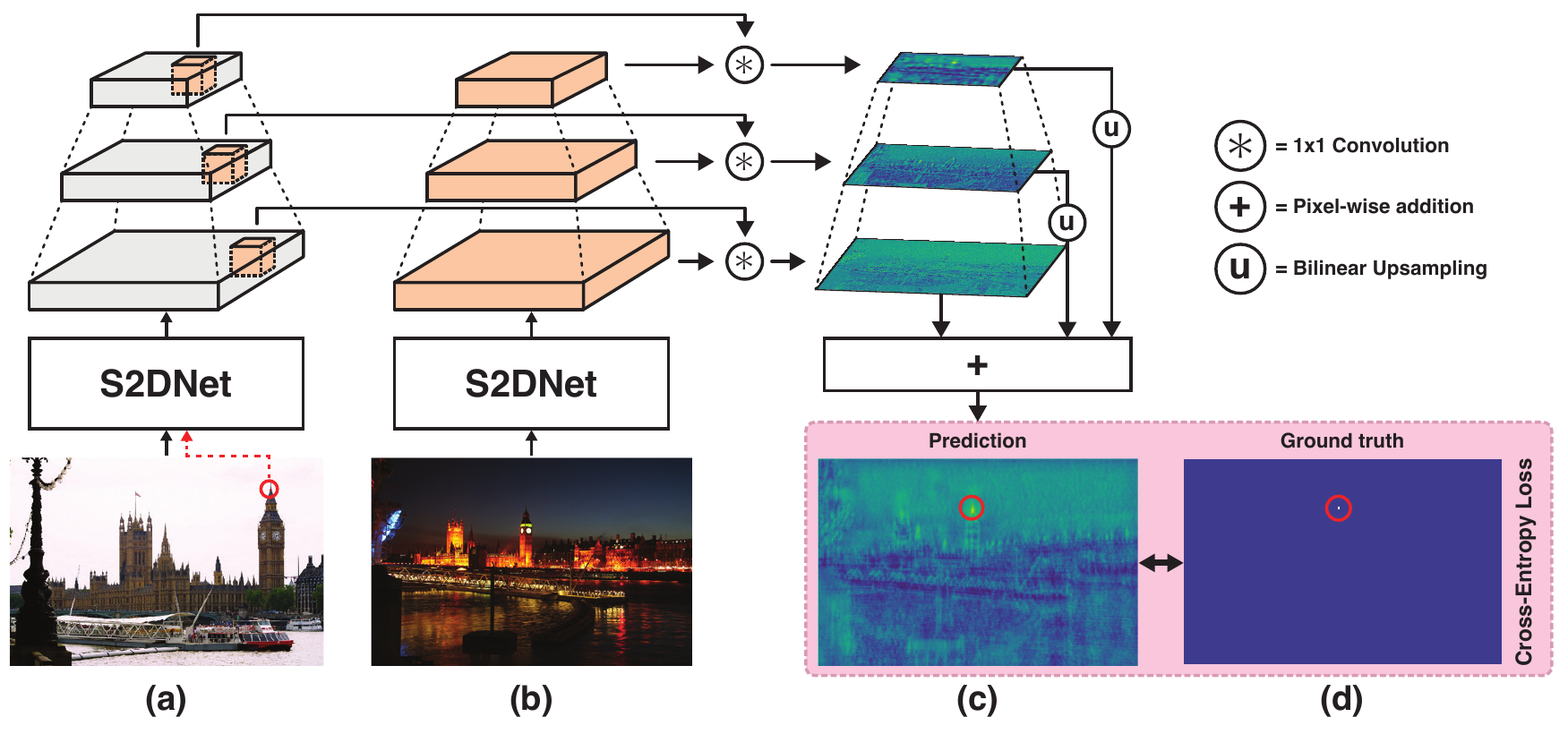}
\end{center}
\caption{\textbf{S2DNet feature matching pipeline overview.}
Given an image and a set of detections coming from an off-the-shelf
keypoint detector $(a)$, we first extract a set of sparse multi-level descriptors with
S2DNet. We then compute dense feature maps for a covisible image $(b)$, and
compute multi-level correspondence maps $(c)$, which we aggregate using bilinear
upsampling and addition. Correspondences can the be retrieved using a simple
argmax operator. 
We explicitly train S2DNet to generate accurate and discriminative correspondence maps
using a supervised classification approach $(d)$.}
\label{fig:S2DNet}
\end{figure}

\section{Related Work}

Establishing 2D to 2D correspondences between images is a key step for many
applications in computer vision, whose performance often directly rely on the
quantity and accuracy of such correspondences~\cite{Gauglitz2011EvaluationOI, tinne2007}.
We can distinguish three categories for obtaining such correspondences, which
either rely on a bilateral, no keypoint or asymmetrical detection stage.\\

\noindent\textbf{Sparse-to-sparse feature matching.} The most popular and studied
approach for feature matching is a two-stage pipeline that first detects
interest point locations and assigns a patch-based descriptor to each of them.
Detection is applied on both images to be matched, and we refer to
these \textit{detect-then-describe} approaches as \textit{sparse-to-sparse}
feature matching methods. To perform keypoint detection, a variety of both
hand-crafted~\cite{SURF, Harris88acombined, SIFT, affinedetectors2} and
learning-based~\cite{LIFT, LFNet} detectors have been developed,
each aiming to detect accurate keypoints in both a repeatable as well as
illumination, scale and affine invariant fashion.
For feature description, methods using histograms of local
gradients~\cite{SURF, BRIEF, SIFT, ORB} or learning-based patch
description~\cite{LIFT, LFNet, L2Net, SOSNet, PNNet} have been widely used.

Yet, when working on long-term scenarios where very strong visual changes can
appear, such methods fail to give reliable
correspondences~\cite{6DOFBenchmark}, motivating the need for data-driven
methods leveraging information beyond patch-level. Among them, end-to-end learning-based
pipelines such as LIFT~\cite{LIFT} propose to jointly learn the
detection and description stages.
Methods like LF-Net~\cite{LFNet} or SuperPoint~\cite{superpoint} learn detection
and description in a self-supervised way, using spatial augmentation of images
through affine transformations.
With D2-Net~\cite{D2Net}, Dusmanu~\etal~showed that a single-branch
CNN could both perform detection and description, in a paradigm referred to as
\textit{detect-and-describe}. Their network is trained in a supervised way with
a contrastive loss on the deep local features, using ground-truth pixel-level
correspondences provided by Structure-from-Motion
reconstructions~\cite{megadepth}.
R2D2~\cite{R2D2} builds on the same paradigm and formulates the learning of
keypoint reliability and repeatability together with the detection and description,
using this time a listwise ranking loss.

In order to preserve the accuracy of their correspondences, both
D2-Net~\cite{D2Net} and R2D2~\cite{R2D2} use dilated convolutions. Still when
looking at feature-matching benchmarks like HPatches~\cite{HPatches}, the
mean matching accuracy at error thresholds of one or two pixels is quite low.
This indicates that their detection stage is often off by a couple of pixels. As
shown in Figure~\ref{fig:intro_figure}, these errors have direct repercussions
on the subsequent localization or reconstruction algorithms.\\

\noindent\textbf{Dense-to-dense feature matching.}
Dense-to-dense matching approaches get rid of the detection stage
altogether by finding mutual nearest neighbors in dense feature maps.
This can be done using densely extracted features from a pre-trained
CNN, combined with guided matching from late layers
to earlier ones~\cite{Taira2018InLocIV}.
NCNet~\cite{NCNet} trains a CNN to search in the 4D space of all possible
correspondences, with the use of 4D convolutions. While they can be trained with
weak supervision, dense-to-dense approaches carry high computational cost and memory
consumption which make them hardly scalable for computer vision applications.
Besides, the quadratic complexity of this approach limits the resolution
of the images being used, resulting in correspondences with low accuracy.\\

\noindent\textbf{Sparse-to-dense feature matching.}
Very recently~\cite{3DV2019} proposed to perform the detection stage
asymmetrically. In such setting,
correspondences are searched exhaustively in the counterpart image, by running for
instance a cross-correlation operation on dense feature maps with a sparse set
of local hypercolumn descriptors. While this exhaustive
search used to be a costly operation, it can now be efficiently computed on
GPUs, using batched $1 \times 1$ convolutions. The main appeal for this approach 
is that under strong visual changes, the need for repeatability in keypoint detection
is alleviated, allowing each pixel to be a detection.
Instead, finding corresponding keypoint locations is left to the dense descriptor map.
Preliminary work on sparse-to-dense matching~\cite{3DV2019} has shown that reusing
intermediate CNN representations trained with weak supervision for the task of image retrieval
can lead to significant gains in performance for visual localization. However,
these features are not explicitly learned for feature matching, and thus fail to
give pixel-accurate correspondences. We will refer to this method as S2DHM, for Sparse-to-Dense Hypercolumn Matching, in the rest of the paper.

In this paper, we propose to explicitly learn correspondence maps for the task of
pixel-level matching.


\section{Method}

In this section we introduce and describe our novel sparse-to-dense feature-matching
pipeline, which we call S2DNet.
%
\begin{figure}[t]
\begin{center}
  \includegraphics[width=0.8\columnwidth]{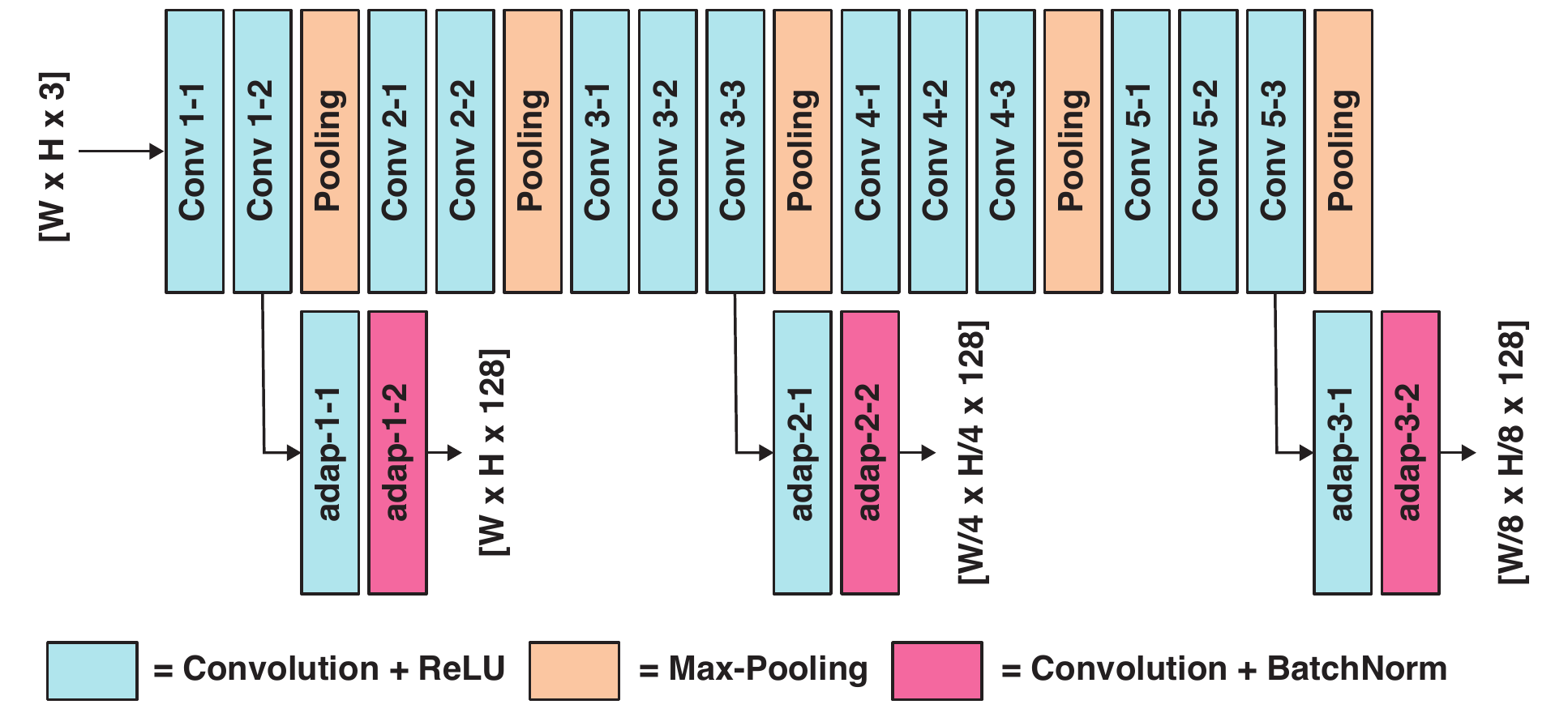}
\end{center}
\caption{\textbf{S2DNet: Architecture overview.} We feed images through a
standard VGG-16~\cite{Simonyan2014VeryDC} backbone, and set three extraction
points to process intermediate features. These features are sent to small,
adaptation layers which help with the convergence and provide more condensed
descriptors.}
\label{fig:architecture}
\end{figure}

\subsection{The Sparse-to-Dense paradigm}
Given an image pair $(\I_\A, \I_\B)$, our goal is to obtain a set of 2D
correspondences which we write as $\{(\p_\A^n, \p_\B^n)\}_{n=1}^N$. 
Let us consider the case where a feature detector (\textit{e.g.} the SuperPoint detector
~\cite{superpoint}) has been applied on image $\A$, producing a set of $N$ keypoints
$\{\p_\A^n\}_{n=1}^N$. In this case, the feature matching problem reduces to a
\emph{sparse-to-dense} matching problem of finding a correspondent $\p_\B^n$ in image 
$\B$ for each detection $\p_\A^n$.
We propose to cast this correspondence learning problem as a supervised classification
task by restricting the set of admissible locations to the pixel coordinates of $\I_\B$.
This leads to the following categorical distribution:

\begin{equation}
 p\left(\p_\B^n|\p_\A^n,\I_\A,\I_\B,\Theta\right)=
 \frac{\exp\left(\C_n[\p_\B^n]\right)}{\sum_{\q\in\Omega}\exp\left(\C_n[\q]\right)} \> ,
\label{eq:likelihood}
\end{equation}
where $\C_n$ is a correspondence map of the size of $\I_\B$ produced by our
novel S2DNet matching pipeline 
and $\Omega$ is the
set of pixel locations of $\I_\B$. S2DNet takes as input $\p_\A^n$, $\I_\A$, $\I_\B$ and its parameters $\Theta$.


\subsection{S2DNet matching pipeline}

We introduce S2DNet, a pipeline built specifically to perform
sparse-to-dense matching which we illustrate in Figure~\ref{fig:S2DNet}. 
Given a pair of images $(\I_\A, \I_\B)$, we apply a convolutional backbone
$\mathcal{F}$ on both images using shared network weights
 \textit{i.e.} $\{\H_\A^m\}_{m=1}^M=\mathcal{F}\left(\I_\A;\Theta\right)$
 and $\{\H_\B^m\}_{m=1}^M=\mathcal{F}\left(\I_\B;\Theta\right)$, where
 $\{\H_\A^m\}_{m=1}^M$ and $\{\H_\B^m\}_{m=1}^M$ correspond to intermediate
 feature maps extracted at multiple levels (see Figure~\ref{fig:architecture}).
 $\Theta$ denotes the parameters of $\mathcal{F}$.
 Such representations are sometimes referred to as
hypercolumns~\cite{Sattler2015HyperpointsAF, Hariharan2014HypercolumnsFO}. While
the earlier layers encode little semantic meaning, they preserve high-frequency
local details which is crucial for retrieving accurate keypoints.
Conversely in the presence of max-pooling, later layers loose in
resolution but benefit from a wider receptive field and thus context.

For each detected keypoint $\p_\A^n$ in $\I_\A$, we extract a set of sparse descriptors in the dense intermediate feature maps 
$\H_\A^m$ and  compute the dense correspondence map $\C_n$ against $\H_\B^m$, by processing each level independently, in the following way:

\begin{equation}
    \C_n = \sum_{m=1}^{M} \mathcal{U}\left(\H_\A^m\left[\p_\A^{n,m}\right] * \H_\B^m\right) \> ,
\label{eq:corresp_map}
\end{equation}
where $\mathcal{U}$ refers to the bilinear upsampling operator to $\I_\B$
resolution, $\p_\A^{n,m}$ corresponds to downscaling the 2D coordinates
$\p_\A^n$ to the resolution of $\H_\A^m$, 
and $*$ is the $1\times1$ convolution operator.


\subsection{Training-time} While state of the art approaches  employ  either a
local  contrastive   or a listwise  ranking loss~\cite{LFNet,  D2Net,  R2D2}  to
train their network, we directly  optimize  for  the  task  of  sparse-to-dense
correspondence retrieval by maximizing the log-likelihood in
eq.(\ref{eq:likelihood}) which results in a  single multi-class cross-entropy
loss.
From a practical point of view, for every training sample, this corresponds to
computing the softmaxed correspondence map and evaluate the cross-entropy loss
using the ground truth correspondence $\p_\B^n$.  This strongly   penalizes
wrong predictions  regardless of  their closeness  to the ground-truth, forces
the network to  generate highly localized and peaked predictions and helps
computing accurate correspondences.

\subsection{Test-time}\label{sec:test-time}
 At test-time, to retrieve the correspondences in $\I_\B$, we proceed as follows
 for each detected keypoint $\p_\A^n$: 
\begin{equation}
  {\p_\B^n}^*=\underset{\p_\B^n}{\text{argmax}} \,\,
p\left(\p_\B^n|\p_\A^n,\I_\A,\I_\B,\Theta\right)=\underset{\p}{\text{argmax}}
\,\, \C_n\left[\p\right] \> ,
\end{equation}
where $\C_n=\text{S2DNet}(\p_\A^n,\I_\A,\I_\B;\Theta).$
By default, S2DNet does not apply any type of filtering and delivers one
correspondence for each detected keypoint in the source image. Since we do not
explicitly deal with co-visibility issues, we filter out some
ambiguous matches if the following condition is not satisfied:
\begin{equation}
  p\left({\p_\B^n}^*|\p_\A^n,\I_\A,\I_\B,\Theta\right)>\tau \> ,
\label{eq:thresholding}
\end{equation}
where $\tau$ is a threshold between $0$ and $1$.

\subsection{S2DNet architecture}
As~\cite{D2Net, 3DV2019}, we use a VGG-16~\cite{Simonyan2014VeryDC} architecture as
our convolutional backbone. We place our intermediate extraction points at three levels, in
\code{conv\_1\_2}, \code{conv\_3\_3} and \code{conv\_5\_3}, after the ReLU
activations. Note that \code{conv\_1\_2} comes before any spatial pooling
layer, and thus preserves the full image resolution.
To both help with the convergence and reduce the final descriptors sizes,
we feed these intermediate tensors to adaptation layers. They consist of two convolutional
layers and a final batch-normalization~\cite{Ioffe2015BatchNA} activation, with an
output size of 128 channels. Refer to Figure~\ref{fig:architecture} for an illustration
of our architecture.

\subsection{Differences with Sparse-to-Dense Hypercolumn
Matching~\cite{3DV2019}} Sparse-to-Dense Hypercolumn Matching
(S2DHM)~\cite{3DV2019} described a weakly supervised approach to learn
hypercolumn descriptors and efficiently obtain correspondences using the
sparse-to-dense paradigm.  In this paper, we propose a supervised alternative
which aims at directly learning accurate correspondence maps. As we will show in
our experiments, this leads to significantly superior results.
Moreover, in its pipeline S2DHM upsamples and concatenates intermediate feature
maps before computing correspondence maps. In comparison, S2DNet computes
correspondence maps at multiple levels before merging the results by addition.
We will later show that the latter approach is much more memory and
computationally efficient.

\section{Experiments} In this section, we evaluate S2DNet on several challenging
benchmarks.  We first evaluate our approach on a commonly used image matching
benchmark, which displays changes in both viewpoint and illumination. We then
evaluate the performance of S2DNet on long-term visual localization tasks, which
display even more severe visual changes.

\subsection{Training data}
We use the same training data as D2-Net~\cite{D2Net} to train S2DNet, which
comes from the MegaDepth dataset~\cite{megadepth}. This dataset consists
of $196$ outdoor scenes and $1,070,568$ images, for which \textit{SfM} was run with
COLMAP~\cite{schoenberger2016sfm, schoenberger2016mvs} to generate a sparse
3D reconstruction. A depth-check is run using the provided depth maps to remove
occluded pixels. As D2-Net, we remove scenes which overlap with the
PhotoTourism~\cite{imc_phototourism,Thomee2016YFCC100MTN} test set.
Compared to D2-Net and to provide strong scale changes, we train S2DNet on
image pairs with an arbitrary overlap.
At each training iteration, we extract random crops of size $512 \times 512$,
and randomly sample a maximum of 128 pixel correspondences. We train S2DNet for
30 epochs using Adam~\cite{Kingma2014AdamAM}. We use an initial learning
rate of $10^{-3}$ and apply a multiplicative decaying factor of $e^{-0.1}$
at every epoch.

\subsection{Image Matching}
We first evaluate our method on the popular image matching benchmark
HPat\-ches~\cite{HPatches}. We use the same 108 sequences of images as
D2-Net~\cite{D2Net}, each sequence consisting of 6 images. These images either
display changes in illumination (for 52 sequences) or changes in viewpoint (for
56 sequences). We consider the first frame of each sequence to be the reference
image to be matched against every other, resulting in 540 pairs of images to
match.\\

\noindent\textbf{Evaluation protocol.} We apply the SuperPoint~\cite{superpoint}
keypoint detector on the first image of each sequence.
For each subsequent pair of images, we perform sparse-to-dense matching using
S2DNet (see section~\ref{sec:test-time}). Additionally, we filter out
correspondences which do not pass the cyclic check of matching back on their
source pixel, which is equivalent to performing a mutual nearest-neighbor
verification as it is done with D2-Net~\cite{D2Net} and R2D2~\cite{R2D2}.

\begin{figure}[t]
\begin{center}
  \includegraphics[width=0.65\columnwidth]{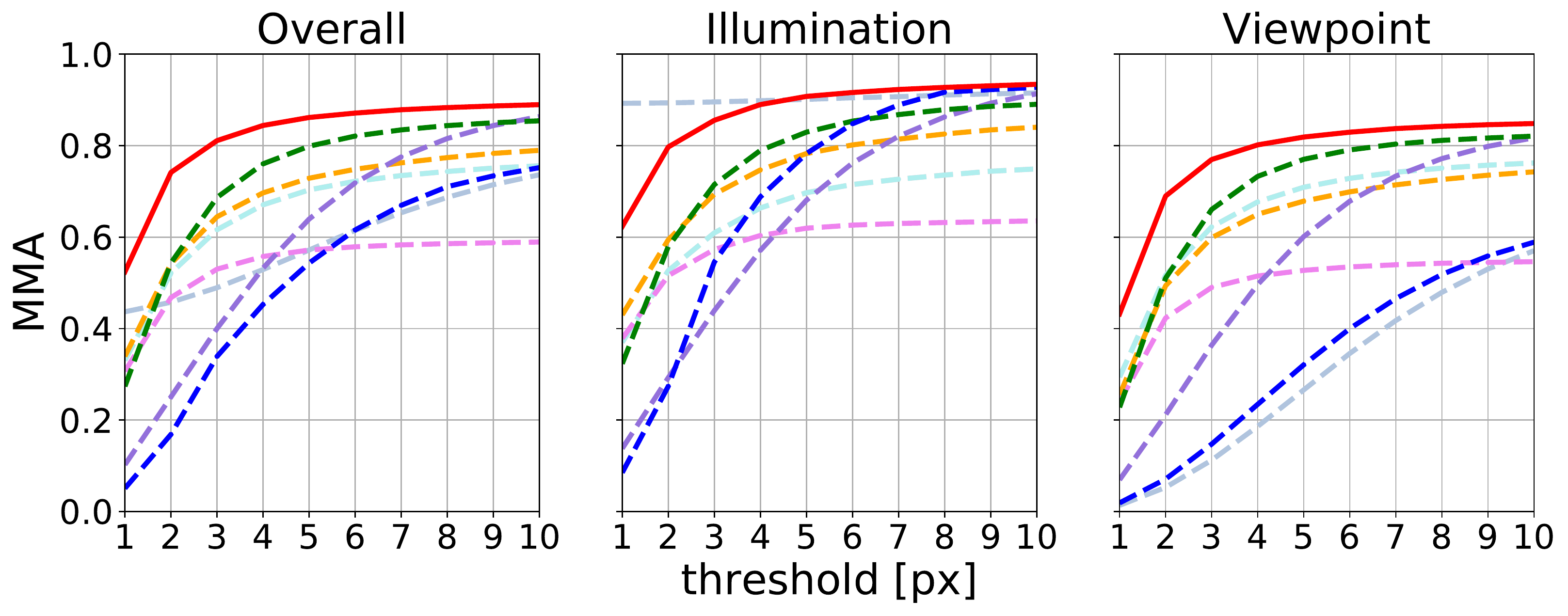}
  \raisebox{0.25\height}{
  \resizebox{0.3\columnwidth}{!}{
    \ra{1.2}
    \begin{tabular}[b]{lcc}\toprule
    \textbf{Method} & \textbf{\#Features} & \textbf{\#Matches} \\ \midrule
      \textcolor{pink}{\rule{0.1cm}{0.6mm} \rule{0.1cm}{0.6mm}} Hes.det.+RootSIFT~\cite{RootSIFT, affinedetectors2} & 6.7K & 2.8K \\
      \textcolor{paleturquoise}{\rule{0.1cm}{0.6mm} \rule{0.1cm}{0.6mm}} HAN+HN++~\cite{HardNet, HesAffRegions} & 3.9K & 2.0K \\
      \textcolor{violet}{\rule{0.1cm}{0.6mm} \rule{0.1cm}{0.6mm}} LF-Net~\cite{LFNet} & 0.5K & 0.2K \\
      \textcolor{orange}{\rule{0.1cm}{0.6mm} \rule{0.1cm}{0.6mm}} SuperPoint~\cite{superpoint} & 1.7K & 0.9K \\
      \textcolor{lightsteelblue}{\rule{0.1cm}{0.6mm} \rule{0.1cm}{0.6mm}} DELF~\cite{DELF} & 4.6K & 1.9K \\
      \textcolor{blue}{\rule{0.1cm}{0.6mm} \rule{0.1cm}{0.6mm}} S2DHM~\cite{3DV2019} & 0.9K & 0.4K\\
      \textcolor{mediumpurple}{\rule{0.1cm}{0.6mm} \rule{0.1cm}{0.6mm}} D2-Net~\cite{D2Net} & 8.3K & 2.8K \\
      \textcolor{lightgreen}{\rule{0.1cm}{0.6mm} \rule{0.1cm}{0.6mm}} R2D2~\cite{R2D2} & 5.0K & 1.8K \\
      \textcolor{red}{\rule{0.3cm}{0.6mm}} S2DNet (ours) & 2.0K & 0.8K \\\bottomrule
    \end{tabular}}}
  \end{center}
  \caption{\textbf{HPatches Mean Matching Accuracy (MMA) comparison.} We report in
  this table the best results for S2DNet, obtained when combined with SuperPoint
  detections. S2DNet outperforms all other baselines, especially at thresholds
  of one or two pixels. This study highlights the power of working in a
sparse-to-dense setting, where every pixel in the target image becomes a
candidate keypoint.}
\label{fig:HPatches}
\end{figure}

We compute the number of matches which fall under multiple reprojection error
thresholds using the ground-truth homographies provided by the dataset, and
report the Mean Matching Accuracy (or MMA) in Figure~\ref{fig:HPatches}.

We compare S2DNet to multiple sparse-to-sparse matching baselines. We report the
performance of RootSIFT~\cite{Arandjelovic2012ThreeTE, RootSIFT} with a Hessian
Affine detector~\cite{affinedetectors2} (Hes.det. + RootSIFT),
HardNet++~\cite{HardNet} coupled with HesAffNet regions~\cite{HesAffRegions}
(HAN + HN++), DELF~\cite{DELF}, LF-Net~\cite{LFNet},
SuperPoint~\cite{superpoint}, D2-Net~\cite{D2Net} and R2D2~\cite{R2D2}.  We also
include results from the sparse-to-dense method S2DHM~\cite{3DV2019}.  \\

\noindent\textbf{Results.}
We find that the best results were achieved when combining
SuperPoint~\cite{superpoint} with a threshold of $\tau=0.20$ (see
Equation~\ref{eq:thresholding}), which are the results reported
in Figure~\ref{fig:HPatches}. We experimentally found that above this
threshold, some sequences obtain very few to no correspondence at all,
which biases the results.
%
%
We show that overall our method outperforms every baselines at any reprojection
threshold. The gain in performance is particularly noticeable at thresholds of 1
and 2 pixels, indicating the correspondences we predict tend to be much more accurate.
DELF~\cite{DELF} achieves competitive results under changes in illumination,
which can be explained by the fact that keypoints are sampled on a fixed grid
and that the images undergo no changes in viewpoint. On the other hand, it
performs poorly under viewpoint changes.
%
\\

\noindent\textbf{Keypoint detector influence.} We run an ablation study
  to evaluate the impact of different feature detectors, confidence thresholds
  as well as using a sparse-to-sparse approach, and report the results in
  Table~\ref{table:ablation} (left).
%
We find that S2DNet tends to work best when combined with
SuperPoint~\cite{superpoint}. We also experimentally find $\tau=0.2$ to be a
good compromise of correspondence rejection while also maintaining a high number
of matches.\\

\noindent\textbf{Sparse-to-sparse vs. Sparse-to-dense.}
%
We find that using S2DNet in a sparse-to-sparse setting (\textit{i.e.} applying
a detector on the image undergoing illumination or viewpoint changes) damages
the results (see Table~\ref{table:ablation}, left). This phenomenon translates
the errors made by keypoint detectors, and motivates the
sparse-to-dense setting. S2DNet efficiently leverages this paradigm and can
find corresponding keypoints that would not have been detected otherwise.
%
Conversely, we study the impact of using sparse-to-sparse learning-based methods
D2-Net~\cite{D2Net} and R2D2~\cite{R2D2} in a sparse-to-dense setting~(see
Table~\ref{table:ablation}, right). We find that using the sparse-to-dense
paradigm systematically improves their performance under illumination changes,
where images are aligned. This suggests that their descriptor maps are robust to
illumination perturbations. On the other hand, performance is damaged for both
methods under viewpoint changes, suggesting that their descriptor maps
are not highly localized and discriminative. Concerning S2DHM, which was
trained in a weakly supervised manner, running it in a sparse-to-sparse setting
improves the accuracy. This highlights the importance of our main contribution,
\textit{i.e.} casting the sparse-to-dense matching problem as a supervised classification
task.

\begin{table}[t]	
  \begin{center}	
    \subfloat[\textbf{S2S vs. S2D - S2DNet descriptors}]{	
    \ra{0.85}	
    \resizebox{0.5\textwidth}{!}{	
      \begin{tabular}{	
          l@{\hskip0.05in}	
          c@{\hskip0.05in}	
          c@{\hskip0.05in}	
          c@{\hskip0.05in}	
          c@{\hskip0.05in}	
          c@{\hskip0.05in}c}%
        \toprule	
        Detector & Matching & $\tau$ & MMA@1 & MMA@2 & MMA@3 & MMA@10 \\ 	
        \midrule	
        & S2D & $0.20$ & \textbf{0.511} &	
        \textbf{0.733} & \textbf{0.805} & \textbf{0.888} \\	
        Harris~\cite{Harris88acombined} & S2D & $0.0$ & 0.441 & 0.626 & 0.690 & 0.787 \\	
        & S2S &  - & 0.278 & 0.464 & 0.565 & 0.763 \\	
        \midrule	
        & S2D & $0.20$ & \textbf{0.511} & \textbf{0.742} &	
        \textbf{0.823} & \textbf{0.902} \\	
        SURF~\cite{SURF} & S2D & $0.0$ & 0.436 & 0.639 & 0.718 & 0.828 \\	
        & S2S & - & 0.302 & 0.506 & 0.619 & 0.829 \\	
        \midrule	
        & S2D & $0.20$ & \textbf{0.487} & \textbf{0.700} &	
        \textbf{0.771} & \textbf{0.851} \\	
        SIFT~\cite{SIFT} & S2D & $0.0$ & 0.441 & 0.626 & 0.690 & 0.787 \\	
        & S2D & - & 0.386 & 0.559 & 0.642 & 0.818 \\	
        \midrule	
        & S2D & $0.20$ & \textbf{0.563} &	
        \textbf{0.747} & \textbf{0.815} & \textbf{0.895} \\	
        SuperPoint~\cite{superpoint} & S2D & $0.0$ & 0.469 & 0.623 & 0.686 & 0.788 \\	
        & S2S & - & 0.373 & 0.599 & 0.709 & 0.847 \\	
        \midrule	
        & S2D & $0.20$ & \textbf{0.467} & \textbf{0.716} &	
        \textbf{0.805} & \textbf{0.911} \\	
        D2-Net~\cite{D2Net} & S2D & $0.0$ & 0.330 & 0.522 & 0.604 & 0.764 \\	
        & S2S & - & 0.118 & 0.285 & 0.425 & 0.777 \\	
        \midrule	
        & S2D & $0.20$ & \textbf{0.478} & \textbf{0.715} &	
        \textbf{0.799} &\textbf{ 0.901} \\	
        R2D2~\cite{R2D2} & S2D & $0.0$ & 0.341 & 0.522 & 0.598 & 0.746 \\	
        & S2S & - & 0.316 & 0.546 & 0.652 & 0.819 \\	
        \bottomrule	
      \end{tabular}%
  }}	
    \subfloat[\textbf{S2S vs. S2D - Other descriptors}]{	
      \raisebox{-0.5\height}{	
      \begin{tabular}[b]{c}	
        \includegraphics[width=0.5\columnwidth]{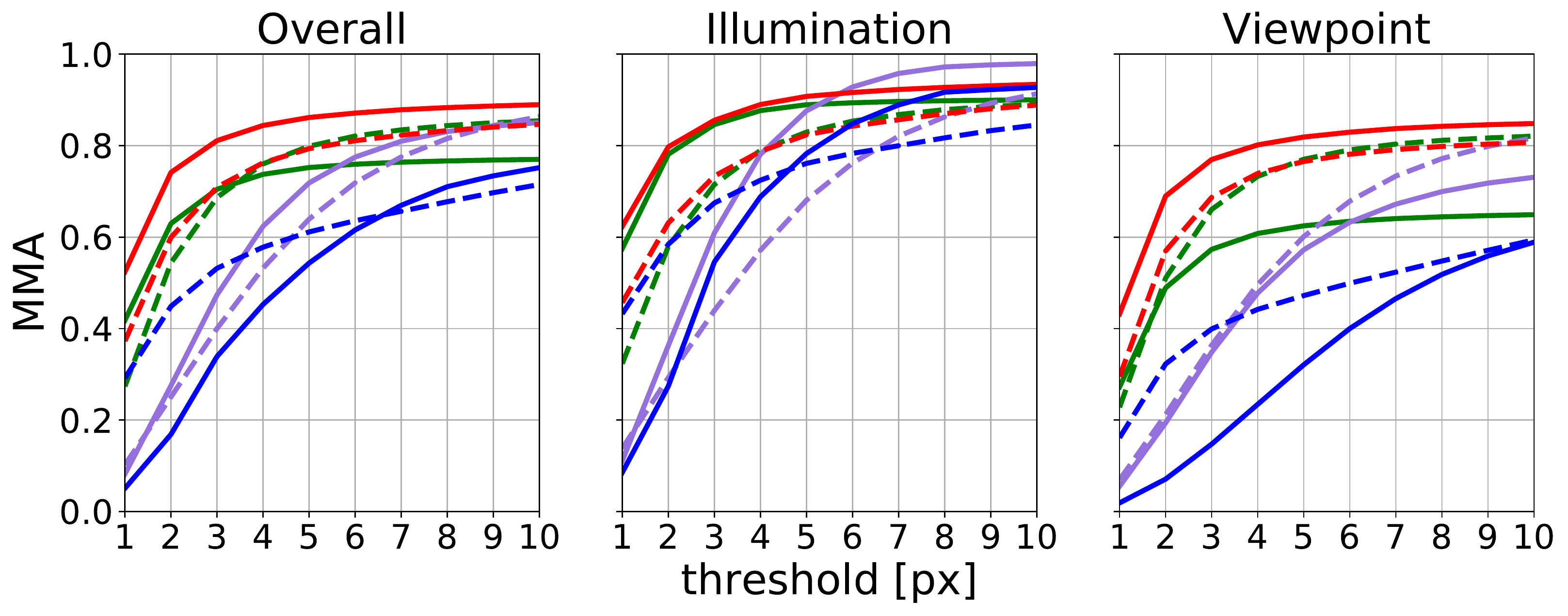}\\	
        \resizebox{0.45\columnwidth}{!}{	
          \ra{1.2}	
          \begin{tabular}[b]{l@{\hskip0.15in}l}	
          \textcolor{mediumpurple}{\rule{0.1cm}{0.6mm} \rule{0.1cm}{0.6mm}}	
          D2-Net (S2S)~\cite{D2Net} &	
          \textcolor{lightgreen}{\rule{0.1cm}{0.6mm} \rule{0.1cm}{0.6mm}}	
          R2D2 (S2S)~\cite{R2D2} \\	
          \textcolor{mediumpurple}{\rule{0.3cm}{0.6mm}}	
          D2-Net (S2D)~\cite{D2Net}&	
          \textcolor{lightgreen}{\rule{0.3cm}{0.6mm}}	
          R2D2 (S2D)~\cite{R2D2}\\	
          \textcolor{blue}{\rule{0.1cm}{0.6mm} \rule{0.1cm}{0.6mm}} 	
          SP + S2DHM~\cite{superpoint, 3DV2019} (S2S) &	
          \textcolor{red}{\rule{0.1cm}{0.6mm} \rule{0.1cm}{0.6mm}} 	
          SP~\cite{superpoint} + S2DNet (S2S) \\	
          \textcolor{blue}{\rule{0.3cm}{0.6mm}} 	
          SP + S2DHM~\cite{superpoint, 3DV2019} (S2D) &	
          \textcolor{red}{\rule{0.3cm}{0.6mm}} 	
          SP~\cite{superpoint} + S2DNet (S2D)\\	
        \end{tabular}}	
    \end{tabular}	
    }}	
    \end{center}	
    \caption{\textbf{Ablation study on HPatches.}	
    In \textbf{(a)}, we evaluate the	
    performance of several detectors in both a sparse-to-dense (S2S) and	
    sparse-to-sparse (S2S) setting using S2DNet descriptors.	
    We find that S2DNet works best in the S2D setting, coupled with SuperPoint	
    (SP)~\cite{superpoint} detections, and a confidence threshold of	
    $\tau=0.20$.	
    In \textbf{(b)}, we study the impact of using sparse-to-sparse learning-based	
    methods in a sparse-to-dense setting. Results lead to the conclusion that	
    D2-Net~\cite{D2Net} and R2D2~\cite{R2D2} descriptor maps are robust to	
  illumination changes but not highly discriminative locally.	
    }\label{table:ablation}	
\end{table}


\subsection{Long-Term Visual Localization}

We showed that S2DNet provides correspondences which are overall more accurate
than other baselines. We will now study its impact for the task of visual
localization under challenging conditions. We report visual localization
results under day-night changes and complex indoor scenes.\\

\begin{figure}[t]
\begin{center}
  \includegraphics[width=1.0\columnwidth]{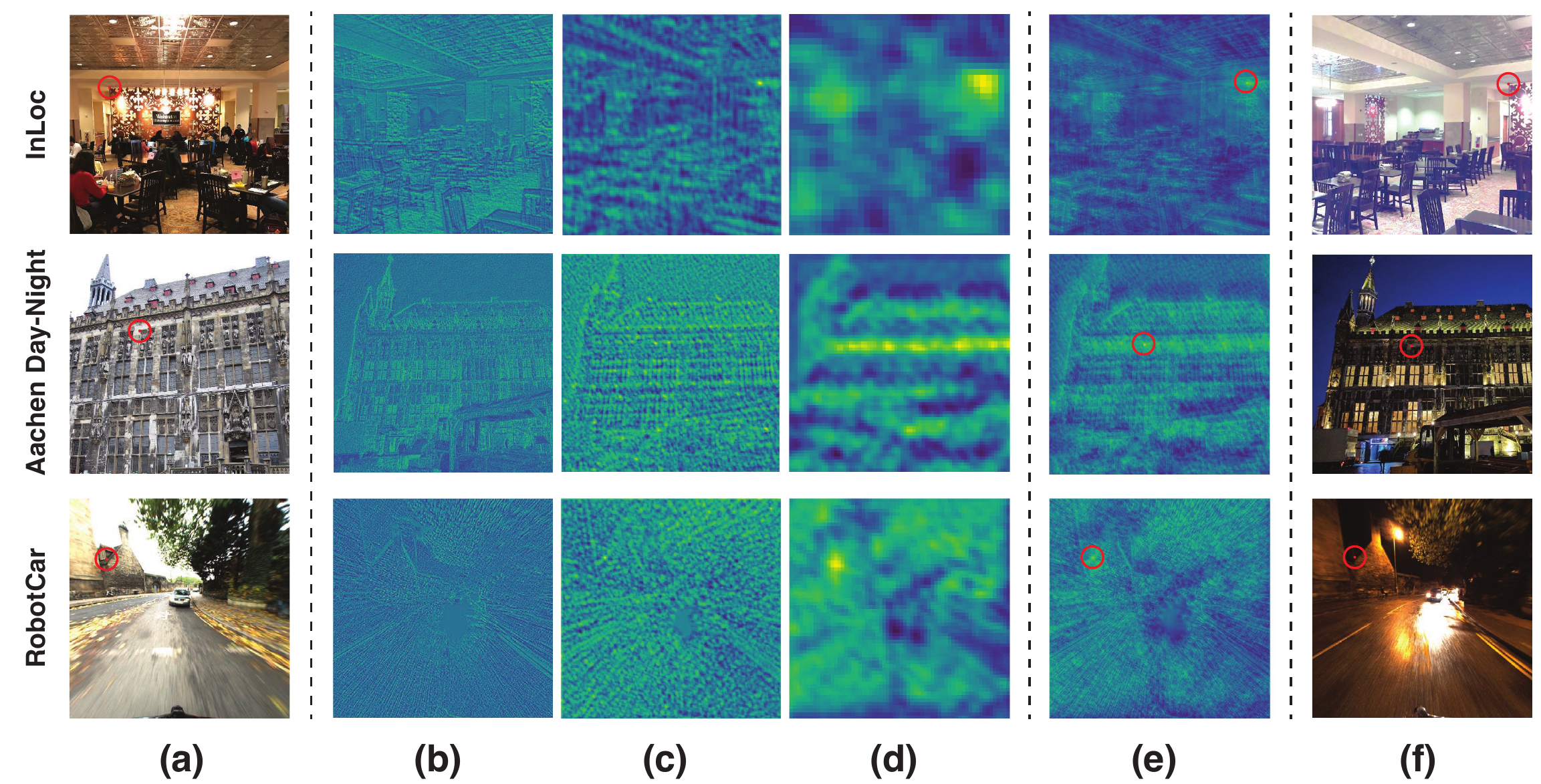}
\end{center}
\caption{\textbf{Correspondence maps examples.} From left to right: Reference
  image with a keypoint detection $(a)$, intermediate correspondence maps
  predicted by S2DNet $(b, c, d)$, aggregated pre-softmax correspondence map
  $(e)$ and retrieved correspondent in the query image $(f)$. From top to
  bottom: images from InLoc~\cite{Taira2018InLocIV}, Aachen
  Day-Night~\cite{Sattler2012ImageRF} and RobotCar Seasons~\cite{Maddern20171Y1}.
}
\label{fig:cmap_qualitative}
\end{figure}

\noindent\textbf{Datasets.}
We evaluate our approach on two challenging outdoor localization datasets which
feature day-to-night changes, and one indoor dataset.
The first dataset is Aachen Day-Night~\cite{Sattler2012ImageRF, 6DOFBenchmark}.
It features 4,328 daytime reference images taken with a handheld smartphone,
for which ground truth camera poses are provided. The dataset also provides a 3D
reconstruction of the scene~\cite{6DOFBenchmark}, built using SIFT~\cite{SIFT}
features and \textit{SfM}. The evaluation is done on 824 daytime and 98 nighttime
images taken in the same environment.
The second dataset is RobotCar Seasons~\cite{Maddern20171Y1}. It features 6,954
daytime reference images taken with a rear-facing camera mounted on a car
driving through Oxford. Similarly, ground truth camera poses and a sparse 3D model
of the world is provided~\cite{6DOFBenchmark} and we localize 3,978 images
captured throughout a year. These images do not only exhibit nighttime
conditions, but also cross-seasonal evolutions such as snow or rain.
Lastly, we evaluate our pipeline on the challenging
InLoc~\cite{Taira2018InLocIV, Wijmans2016Exploiting2F}
dataset. This indoor dataset is difficult because of its large scale,
illumination and long-term changes as well as the presence of repetitive
patterns such as corridors (see Figure~\ref{fig:cmap_qualitative}).
It contains $9,972$ database and $356$
high-resolution query images, as well as dense depth maps which can be used to
perform dense pose verification.
We report for each datasets the pose recall at three position and orientation
thresholds for daytime and nighttime query images, as per~\cite{6DOFBenchmark}. \\
\begin{table}[t]
  \ra{1.05}
  \begin{center}

    \resizebox{0.6\textwidth}{!}{
      \begin{tabular}{@{} ll *{4}{c@{\hskip0.1in}} @{}}%
        \multicolumn{1}{c}{}
        & \multicolumn{3}{c}{\makecell{
        \textit{InLoc}\textit{(fixed pipeline)}}}\\
        \toprule
        \multicolumn{1}{l}{Method}
        & \multicolumn{3}{c}{Threshold Accuracy}\\
        \cmidrule(lr){2-4}
        &\makecell{0.25m \\ 2\degree}
        & \makecell{0.5m \\ 5\degree}
        & \makecell{5m \\ 10\degree}\\
        \midrule
        Direct PE - Aff. RootSIFT~\cite{affinedetectors2, RootSIFT} & 18.5 & 26.4 & 30.4\\
        Direct PE - D2-Net~\cite{D2Net} & 27.7 & 40.4 & \textbf{48.6}\\
        \textbf{Direct PE - S2DNet (ours)} & \textbf{29.3} & \textbf{40.9} & 48.5\\
        \midrule
        Sparse PE - Aff. RootSIFT~\cite{affinedetectors2, RootSIFT} & 21.3 & 32.2 & 44.1\\
        Sparse PE - D2-Net~\cite{D2Net} & 35.0 & 48.6 & 62.6 \\
        \textbf{Sparse PE - S2DNet (ours)} & \textbf{35.9} & \textbf{49.0} &
        \textbf{63.1}\\
        \midrule
        Sparse PE + Dense PV - Aff. RootSIFT~\cite{affinedetectors2, RootSIFT}
                                           & 29.5 & 42.6 & 54.5 \\
        Sparse PE + Dense PV - D2-Net~\cite{D2Net}
                                           & 38.0 & \textbf{56.5} & 65.4 \\
        \textbf{Sparse PE + Dense PV - S2DNet (ours)}
                      & \textbf{39.4} & 53.5 & \textbf{67.2}\\
        \midrule
        \gray{Dense PE + Dense PV - InLoc}~\cite{Taira2018InLocIV}
                      & \gray{38.9} & \gray{56.5} & \gray{69.9} \\
        \bottomrule
      \end{tabular}%
    }
    \quad
    \resizebox{0.3\textwidth}{!}{
      \begin{tabular}{@{} ll *{4}{c@{\hskip0.1in}} @{}}%
        \multicolumn{1}{c}{}
        & \multicolumn{3}{c}{\makecell{\textit{Aachen Day-Night}
        \\\textit{(fixed pipeline)}}}\\
        \toprule
        \multicolumn{1}{l}{Method}
        & \multicolumn{3}{c}{Threshold Accuracy}\\
        \cmidrule(lr){2-4}
        &\makecell{0.25m \\ 2\degree}
        & \makecell{0.5m \\ 5\degree}
        & \makecell{5m \\ 10\degree}\\
        \midrule
        RootSIFT~\cite{RootSIFT} & 3.7 & 52.0 & 65.3 \\
        HAN+HN~\cite{HesAffRegions} & 37.8 & 54.1 & 75.5\\
        SuperPoint~\cite{superpoint} & 42.8 & 57.1 & 75.5\\
        DELF~\cite{DELF} & 39.8 & 61.2 & 85.7\\
        D2-Net~\cite{D2Net} & 44.9 & 66.3 & \textbf{88.8}\\
        R2D2~\cite{R2D2}* & \textbf{45.9} & 66.3 & \textbf{88.8}\\
        S2DNet (ours) & \textbf{45.9} & \textbf{68.4} & \textbf{88.8}\\
        \bottomrule
      \end{tabular}%
    }
    \quad
  \end{center}
  \caption{\textbf{InLoc~\cite{Taira2018InLocIV} (left) and Local Features
      Benchmark~\cite{6DOFBenchmark} (right) results}.
    We report localization recalls in percent, for three translation and orientation
    thresholds.
    On InLoc, S2DNet outperforms both baselines at the finest threshold for the
    sparse categories. We also include Dense PE baseline results for reference.    
    R2D2 authors did not provide results on this benchmark.
    %
    On the local features benchmark (a pre-defined localization pipeline),
    S2DNet achieves state-of-the-art results at the medium precision threshold.
    Due to the relatively small number of query images however, recent methods
    like D2-Net and R2D2 are saturating around the same performance.
    *Note that R2D2 was trained on Aachen database images.
    }
 \label{table:LFC_results}\label{table:inloc_results}
\end{table}

\noindent\textbf{Indoor Localization.}
The InLoc~\cite{Taira2018InLocIV} localization benchmark comes with a
pre-defined code base and several pipelines for localization.
%
%
The first one is called Direct Pose Estimation (Direct PE) and
performs hierarchical localization using the set of top-ranked database images
obtained using image retrieval, followed by
P3P-LO-RANSAC~\cite{Lebeda2012FixingTL,Fischler1981RandomSC}.  The second
variant applies an intermediate spatial verification
step~\cite{Philbin2007ObjectRW} to reject outliers, referred to as (Sparse PE).
On top of this second variant, Dense Pose Verification (Dense PV) can be applied
to re-rank pose candidates by using densely extracted RootSIFT~\cite{RootSIFT}
features.
In each variant, we use S2DNet to generate 2D-2D correspondences between queries
and database images, which are then converted to 2D-3D correspondences using the
provided dense depth maps. We use a SuperPoint~\cite{superpoint} detector and
mutual nearest-neighbour filtering.

InLoc localization results are reported in Table~\ref{table:inloc_results}.
We compare our approach to the original InLoc baseline which uses
affine covariant~\cite{affinedetectors2} detections and RootSIFT~\cite{RootSIFT}
descriptors, as well as results provided by D2-Net~\cite{D2Net}. 
%
We find that S2DNet outperforms both sparse baselines at the finest threshold, and
is on par with other methods at the medium and coarse thresholds.  In the sparse
setting, best results are achieved when combined with geometrical and dense pose
verification (Sparse PE + Dense PV).  In addition we include localization
results that were computed by the benchmark authors using dense-to-dense feature
matching (Dense PE). Due to the nature of our pipeline and the very high
  memory and computational consumption of this variant, we choose to limit our
  study to sparse correspondence methods. It is interesting to note however that
  S2DNet outperforms the original (Dense PE + Dense PV) InLoc baseline at the
finest precision threshold, using a much lighter computation.\\

\begin{table}[t]
  \ra{1.1}
  \begin{center}
    \resizebox{\textwidth}{!}{
      \begin{tabular}{@{} ll *{14}{c@{\hskip0.1in}} @{}}%
        & \multicolumn{1}{c}{}
        & \multicolumn{6}{c}{\textit{RobotCar Seasons}}
        & \multicolumn{6}{c}{\textit{Aachen Day-Night}}\\
        \cmidrule(lr){3-8}
        \cmidrule(lr){9-14}
        \multicolumn{2}{c}{}
        & \multicolumn{3}{c}{\textit{Day-All}}
        & \multicolumn{3}{c}{\textit{Night-All}}
        & \multicolumn{3}{c}{\textit{Day}}
        & \multicolumn{3}{c}{\textit{Night}}\\
        \toprule
        \multicolumn{2}{c}{Method}
        & \multicolumn{3}{c}{Threshold Accuracy}
        & \multicolumn{3}{c}{Threshold Accuracy}
        & \multicolumn{3}{c}{Threshold Accuracy}
        & \multicolumn{3}{c}{Threshold Accuracy}\\
        \cmidrule(lr){3-5} \cmidrule(lr){6-8}
        \cmidrule(lr){9-11} \cmidrule(lr){12-14} &
        &\makecell{0.25m \\ 2\degree} & \makecell{0.5m \\ 5\degree} & \makecell{5m \\ 10\degree}
        &\makecell{0.25m \\ 2\degree} & \makecell{0.5m \\ 5\degree} & \makecell{5m \\ 10\degree}
        &\makecell{0.25m \\ 2\degree} & \makecell{0.5m \\ 5\degree} & \makecell{5m \\ 10\degree}
        &\makecell{0.25m \\ 2\degree} & \makecell{0.5m \\ 5\degree} &
        \makecell{5m \\ 10\degree}\\
        \midrule

        \parbox[t]{8mm}{\multirow{3}{*}{\rotatebox[origin=c]{90}{
            \scriptsize{\makecell{Structure\\-based}}}}} &
        CSL~\cite{CSL} & 45.3 & 73.5 & 90.1
                       & 0.6 & 2.6 & 7.2
                       & 52.3 & 80.0 & 94.3
                       & 24.5 & 33.7 & 49.0\\
        & AS~\cite{ActiveSearch} & 35.6 & 67.9 & 90.4
                                 & 0.9 & 2.1 & 4.3
                                 & 57.3 & 83.7 & \underline{96.6} &
                                 19.4 & 30.6 & 43.9\\
        & SMC~\cite{SMC}~* & \gray{50.3} &
              \gray{79.3} & \gray{95.2} & \gray{7.1} &
              \gray{22.4} & \gray{45.3}
              & - & - & - & - & - & -\\
        \hline

        \parbox[t]{8mm}{\multirow{3}{*}{\rotatebox[origin=c]{90}{
                           \scriptsize{\makecell{Retrieval\\-based}}}}} &
        FAB-MAP~\cite{Cummins2008FABMAPPL} & 2.7 & 11.8 & 37.3
                                           & 0.0 & 0.0 & 0.0
                                           & 0.0 & 0.0 & 4.6
                                           &0.0 & 0.0 & 0.0\\
        & NetVLAD~\cite{NetVLAD} & 6.4 & 26.3 & 90.9
                                 & 0.3 & 2.3 & 15.9
                                 &0.0 & 0.2 & 18.9
                                 & 0.0 & 2.0 & 12.2\\
        & DenseVLAD~\cite{DenseVLAD} & 7.6 & 31.2 & 91.2
                                     & 1.0 & 4.4 & 22.7
                                     & 0.0 & 0.1 & 22.8
                                     & 0.0 & 2.0 & 14.3\\
        \hline
        \parbox[t]{8mm}{\multirow{4}{*}{\rotatebox[origin=c]{90}{
            \scriptsize{\makecell{Hierar\\-chical}}}}}
        & HF-Net~\cite{Sarlin2019FromCT} & 53.0 & 79.3 & 95.0
                           & 5.9 & 17.1 & 29.4
                           & 79.9 & 88.0 & 93.4
                           & 40.8 & 56.1 & 74.5\\
        & S2DHM~\cite{3DV2019}~* & \gray{45.7} & \gray{78.0} & \gray{95.1}
                           & \gray{22.3} & \gray{61.8} & \gray{94.5}
                           & 56.3 & 72.9 & 90.9
                           & 30.6 & 56.1 & 78.6\\
        & D2-Net~\cite{D2Net} & \textbf{54.5} &
            \underline{80.0} & \underline{95.3}
                          & \textbf{20.4} & \underline{40.1} & \underline{55.0}
                          & \textbf{84.8} & \textbf{92.6} & \textbf{97.5}
                          & \underline{43.9} & \underline{66.3} & \underline{85.7}\\
        & \textbf{S2DNet (ours)} & \underline{53.9} & \textbf{80.6} & \textbf{95.8}
                          & \underline{14.5} & \textbf{40.2} & \textbf{69.7}
                          & \underline{84.3} & \underline{90.9} & 95.9
                          & \textbf{46.9} & \textbf{69.4} & \textbf{86.7}\\
        \bottomrule
      \end{tabular}%
    }
  \end{center}
  \caption{\textbf{Localization results}.
    We report localization recalls in percent, for three translation and orientation
    thresholds (\textit{high}, \textit{medium},  and \textit{coarse}) as in \cite{6DOFBenchmark}.
    We put in bold the \textbf{best} and underline the \underline{second-best} performances
    for each threshold.
    S2DNet outperforms every baseline in nighttime conditions, except at the
    finest threshold of RobotCar Seasons. This can be explained by the extreme
    visual changes and blurriness that these images undergo. At daytime, S2DNet
    performance is on par with D2-Net~\cite{D2Net}.
    *Note that S2DHM~\cite{3DV2019} was trained directly on RobotCar
    sequences, which explains the high nighttime performance. SMC~\cite{SMC}
    also uses additional semantic data and assumptions. R2D2~\cite{R2D2} authors did not
  provide localization results on these benchmarks.}
    \label{table:localization_results}
\end{table}

\noindent\textbf{Day-Night Localization.} 
We report day-night localization results with S2DNet using two localization
protocols.
%
%
Localization results reported in Table~\ref{table:LFC_results}
show that S2DNet achieves state-of-the-art results, outperforming all other
methods at the medium precision threshold. 
It is important to note that R2D2 was finetuned on Aachen database images.

We then report in Table~\ref{table:localization_results} localization results
using a hierarchical approach, similar to~\cite{Sarlin2019FromCT, 3DV2019,
6DOFBenchmark}. Contrary to Table~\ref{table:inloc_results}, these results do
not allow to compare the keypoint matching approaches alone since localization
pipelines are different. Even the comparison with D2Net is difficult to
interpret since their full localization pipeline was not released. Still, S2DNet
achieves state-of-the-art results in Aachen nighttime images, and outperforms
all baselines that were not trained on RobotCar nighttime images at medium and
coarse precision thresholds. At daytime, where detecting repeatable and accurate
keypoints is easier, S2DNet is on par with other learning-based methods. At the
finest nighttime RobotCar threshold it is likely that S2DNet features struggle
to compute accurate correspondences, which can be explained by the extreme
visual changes these images undergo (see Figure~\ref{fig:cmap_qualitative}).
Overall, this study shows that S2DNet
achieves better performance in particularly challenging conditions such as
nighttime, compared to other sparse-to-sparse alternatives.\\

\begin{table}[t]
  \ra{1.0}
  \begin{center}
    \resizebox{1.0\textwidth}{!}{
      \begin{tabular}{
          cc@{\hskip 0.1in}
          c@{\hskip 0.1in}
          c@{\hskip 0.1in}
          c@{\hskip 0.1in}
          c@{\hskip 0.1in}
          c@{\hskip 0.1in}
          c@{\hskip 0.1in}c}%
        \toprule
        Method
               & \makecell{Network\\Backbone}
               & \makecell{Descriptor\\Size}
               & \makecell{Forward \\pass on $\I_q$}
               & \makecell{Detection \\ step on $\I_q$}
               & \makecell{Matching\\per keypoint}
               & \multicolumn{3}{c}{\makecell{Total online\\computational time} }\\
        \midrule
        & & & $t_A$ & $t_B$ & $t_C$ & \multicolumn{3}{c}{$t_A + t_B + N \times K \times t_C$}\\
        \midrule
                & & & & & & \makecell{$N=1$\\$K=1000$} & \makecell{$N=5$\\$K=1000$} & \makecell{$N=15$\\$K=1000$}\\
        \midrule
        D2-Net~\cite{D2Net} & VGG-16 & $512$ & 17.8ms & 5,474s &
         0.4$\mu$s & 5,492s & 5,494s & 5,498s\\
        R2D2~\cite{R2D2} & L2-Net & $128$ 
                         & 19.1ms & 479.6ms & 0.2$\mu$s & 0,499s & \textbf{0,499s} & \textbf{0,501s}\\
        S2DHM~\cite{3DV2019} & VGG-16
                             & $2048$ & 326ms & -
                             & 0.33ms
                             & 0,656s & 1,976s & 5,276s\\
        \midrule
        S2DNet & \makecell{VGG-16 + adap.}
               & $3 \times 128$ & 28.2ms
               & - & 0.31ms
               & \textbf{0,338s} & 1,578s & 4,678s\\
        \bottomrule
      \end{tabular}%
    }
  \end{center}
  \caption{\textbf{Computational Time Study for visual localization.} We compare
    the time performance of our method against other learning-based approaches
    in a visual localization scenario.
    For a given query image $\I_q$ and $N$ reference
    images of size $1200 \times 1600$ with $K$ detections each, we report the average
    measured time to perform image matching against each of them.
    In this standard setting, keypoint
    locations and descriptors have already been extracted offline from the
    reference images.
  }
\label{table:time_perf}
\end{table}

\section{Discussion}
\noindent\textbf{Runtime performance.} To compare S2DNet against
state-of-the-art approaches, we time its performance for the scenario of visual
localization.
We run our experiments on a machine equipped with an Intel(R) Xeon(R) E5-2630
CPU at 2.20GHz, and an NVIDIA GeForce GTX 1080Ti GPU. We report the results in
Table~\ref{table:time_perf}.
In a localization setting, we consider the keypoint detection and description
step to be pre-computed offline for reference images.  Thus for an incoming
query image, only sparse-to-sparse methods need to perform the keypoint
detection and descriptor extraction step.  We find this very step to be the
bottleneck of learning-based methods like D2-Net~\cite{D2Net} or
R2D2~\cite{R2D2}. Indeed, these methods are slowed down by the non-maxima
suppression operations, which are in addition run on images of multiple scales. 
For S2DHM~\cite{3DV2019} and S2DNet, no keypoint detection is performed on the
incoming query image and most of the computation lies in the keypoint matching
step.
As expected however, the matching step is much more costly for these
sparse-to-dense methods. Still,  for 1000 detections and 1 retrieved image, S2DNet is the fastest method while for
15 retrieved image, it is on par with D2-Net.\\

\noindent\textbf{Current limitations of the sparse-to-dense paradigm.}
One limitation of our current sparse-to-dense matching formulation appears
for the task of multiview 3D reconstruction. Indeed, the standard approach to
obtain features tracks consists in 1) detecting and describing keypoints in
each image, 2) matching pairs of images using the previously extracted
keypoints descriptors and 3) creating tracks from these matches. In our S2D
matching paradigm, every pixel becomes a detection candidate which is not
compatible with the standard 3D reconstruction pipeline previously described.
This limitation opens novel directions of research for rethinking the standard
tracks creation pipeline and enabling the use of S2D matching in 3D
reconstruction frameworks.

\section{Conclusion} In this paper we presented S2DNet, a new sparse-to-dense
learning-based keypoint matching architecture. In contrast to other
sparse-to-sparse methods we showed that this novel pipeline achieves superior
performance in terms of accuracy, which helps improve subsequent long-term
visual localization tasks.  Under visually challenging conditions, S2DNet
reaches state-of-the-art performance for image matching and localization, and
advocates for the development of sparse-to-dense methods.


\section*{Acknowledgement}
This project has received funding from the Bosch Research Foundation~(\emph{Bosch Forschungsstiftung}).
We gratefully acknowledge the support of NVIDIA Corporation with the donation of the Titan Xp GPU used for this research.


\bibliographystyle{splncs04}
\bibliography{string,bibliography}

\end{document}